# Critical Evaluation of LOCO dataset with Machine Learning

*Recep Savas, Johannes Hinckeldeyn*

Hamburg, University of Technology, Institute for Technical Logistics, Theodor-Yorck-Straße 8, 21079 Hamburg, e-mail: itl@tuhh.de

**Purpose:** *Object detection is rapidly evolving through machine learning technology in automation systems. Well prepared data is necessary to train the algorithms. Accordingly, the objective of this paper is to describe a re-evaluation of the so-called Logistics Objects in Context (LOCO) dataset, which is the first dataset for object detection in the field of intralogistics.*

**Methodology:** *We use an experimental research approach with three steps to evaluate the LOCO dataset. Firstly, the images on GitHub were analyzed to understand the dataset better. Secondly, Google Drive Cloud was used for training purposes to revisit the algorithmic implementation and training. Lastly, the LOCO dataset was examined, if it is possible to achieve the same training results in comparison to the original publications.*

**Findings:** *The mean average precision, a common benchmark in object detection, achieved in our study was 64.54%, and shows a significant increase from the initial study of the LOCO authors, achieving 41%. However, improvement potential is seen specifically within object types of forklifts and pallet truck.*

**Originality:** *This paper presents the first critical replication study of the LOCO dataset for object detection in intralogistics. It shows that the training with better hyperparameters* *based on LOCO can even achieve a higher accuracy than presented in the original publication. However, there is also further room for improving the LOCO dataset.*

# 1    Introduction

Object detection is a rapidly evolving through machine learning technology, which is one of the most important computer vision applications. It allows to detect, identify, and track objects in images and videos by using advanced level of neural network algorithms (Benjdira, Khursheed and Koubaa, 2018). Object detection, unlike object recognition, involves finding the coordinates of the detected object on the image. With the coordinates found from the detection information, the object is determined in the area where it will be enclosed with a bounding box.

There are two types of approaches to create neural networks for detecting objects on images. First of them are two stage object detectors which break down the detection application by means of identifying object regions and then classifying the image within its region to determine object classes. The popular two stage algorithms, such as Faster R-CNN, use a Region Proposal Network (Zhao, Xu and Wu, 2019), which simultaneously makes a prediction for object bounds and objectness scores at each position in the image. Since there are multiple iterations running at the same time within the algorithm, it slows down the detection speed in the application and makes it inconvenient for real time detection in videos.

Secondly, the other approach to create object detection neural network is called one stage algorithm. You Only Look Once (YOLO) is one of the best examples for one stage algorithms in object detection (Bochovskiy, 2020). It can process the image so fast that it predicts the class and coordinates of all objects in the image by passing the image through the neural network at only one time. The basis of this estimation process lies in the fact that it treats object detection as a single regression problem. The algorithm first splits the input image into SxS grids. These grids can be divided into 5x5, 9x9 or 21x21 tiles. Each grid in the image is responsible to figure out, whether there is any object within its individual region. If there is an object detected, it will recognize the classes with length, height information for bounding box information. This formulation functionality makes YOLO perfect in object detection. That is why, YOLO is also preferred in our study.

Logistics applications are one of the most suitable areas where object detection technology can be adapted (Li et al., 2019). The so-called Logistics Objects in COntext (LOCO) dataset originates from the Technical University of Munich and is the first available open source dataset with relation to intralogistics (Mayershofer et al., 2020). It was created to detect five different types of objects: pallets, small load carriers, stillages, forklifts and pallet trucks. It contains a total of 151,427 annotations from 5,097 images. The dataset is available on GitHub (GitHub, 2022a) and described in the paper from Mayershofer et al.. A critical examination of the content of the dataset and its potential to train algorithms for objection detection has not been available so far. Therefore, the objective of this paper is to describe a re-evaluation of the LOCO dataset and its capabilities.

## 2    Related work

There are significant studies accomplished in artificial intelligence field about re-evaluating the existing datasets by using different evaluation parameters and setting high variety of performance indicators (Pipino et al., 2002). Keeping the main evaluation parameters of the origin of the related datasets is suggested in most of the re-evaluation research. Moreover, it is also advised to include some additional performance indicators or bringing own evaluation methodology (Al-Riyami et al., 2018), depending on the machine learning application. Mayershofer et al. train the LOCO dataset by using YOLO and Faster R-CNN algorithms and evaluated them by using mean average precision (mAP)@50 results (Bochovskiy, 2020). Within our studies, we also kept the mAP@50 parameter as performance indicator. At the same time, to check the responds of the YOLO algorithm in different object types, we used highly diverse test images (about 500) from online platforms. Furthermore, since object detection algorithms are mostly used in real time video detections, it was important part of our study to examine the performance of the LOCO dataset with YOLO algorithms. Mayershofer et al. mentioned in its publications that they tested the performance of the LOCO dataset in real time.

However, testing experiences about the real time performance of the LOCO dataset are not available in GitHub.

Tuning hyperparameters is mostly suggested in re-evaluating a dataset (Nematzadeh et al., 2022). Tuning has an important role to improve the accuracy of the evaluations. Because of that, we made over 35 different trainings by changing hyperparameters, until we find the best weights of our YOLO model.

Moreover, even though the state of art neural network is used for object detection application, it is always possible that there might be a chance of improvement in detection performance. The training datasets are one of the strongest pillars that sustain machine learning algorithms. If the desired output for object detection is not attained, some further techniques should be accomplished on the dataset to improve its positive results. In the previous studies, it was experienced that the increasing the amount of data artificially, improves the detection results augmentation (Apple Developer Documentation, 2022) can be used by enhancing the dataset artificially. The dataset is an element that affects the efficiency in the object detection application considerably. There are two possible methods to increase the amount of data artificially: data augmentation and preparation of synthetic data. Within our study to increase the number of data artificially, we used data augmentation techniques. (Sessions and Valtorta, 2006).

## 3    Methodology

We use a research approach with three steps to re-evaluate the LOCO dataset. Firstly, the images on GitHub were prepared for the training and analyzed for a deeper understanding of the dataset. The json file called "loco-all-v1.json" is already shared including all labels for each annotated image. However, annotations in this file are stored in COCO format (COCO - Common Objects in Context, 2022). Therefore, the data is converted from COCO format to YOLO format in the following stages of the data preprocessing. At the same time, since the data is too detailed for our object detection application, the required information for our training should have been sorted out. That is why, at the beginning, the necessary columns for each annotation are reduced by using pandas library to order the data properly.

Training in YOLO needs a txt file for each image with classification and localization information of objects in the image. Localization information is given within the bounding box coordinate system in txt files. There are different mathematical coordinates for each COCO and YOLO formats in bounding box representations (13.3. Object Detection and Bounding Boxes — Dive into Deep Learning 0.17.5 documentation, 2022) and bounding boxes for each object had to be converted. Table 1 shows, how the bounding box coordinates differ between COCO and YOLO formats .

Table 1: Bounding Box Orientations for COCO and YOLO Formats

| Format | Arrow | Example |
|--------|-------|---------|
| COCO | [x_min, y_min, x_max, y_max] | [98, 345, 420, 462] |
| YOLO | N[x_min, y_min, x_max, y_max] | [0.153125, 0.71875, 0.65625, 0.9625] |

Secondly, after the dataset was ready to train in YOLO format, an appropriate method had to be chosen to reach high processing speed to reduce the training durations. A workspace in Google Drive was created and Colab Pro+ was used to have priority access to Google's Graphic Processing Units (GPU) to train with the LOCO dataset by using YOLOv4 and YOLOv4-tiny algorithms. Most of the time NVIDIA Tesla V100 was assigned to us by Google servers. The biggest advantage of choosing YOLO algorithms is its superb speed. it is excessively faster than Faster R-CNN in detection speed where the frame per second (FPS), while giving higher mean average precisions, as shown in the Figure 1 (Kim et al., 2020).

Critical Evaluation of LOCO dataset with Machine Learning

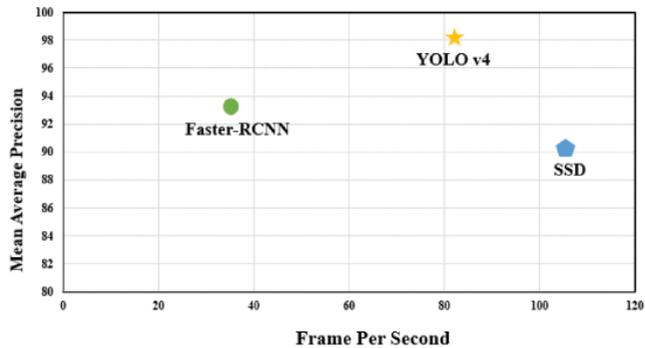

Figure 1: Faster R-CNN and YOLOv4 comparison

On the last part of our study, we focused on finding answers for our research questions. First question of the project is to examine whether it is possible to replicate the findings from LOCO Dataset according to its publications and the information on their GitHub repository. The second research question, which is to define the strengths and weaknesses of LOCO Dataset, is examined by using the high number of trainings with variously tuned hyperparameters. The algorithm was also tested with a wide range of images from a test set to examine whether the dataset gives satisfactory results in different kind of intralogistics environments. Finally, as third research question, some data augmentation techniques on object detection are examined in the LOCO dataset.

1 Examining the reproducibility of LOCO dataset

2 Strengths and weaknesses of the LOCO dataset

3 Data augmentation of the LOCO dataset

Figure 2: List of research questions

At the end of our studies, all these questions are answered objectively to increase the overall performance of the LOCO dataset for better object detection in the field of intralogistics. To be able to find some answers for our questions raised while detecting forklifts and pallet trucks, we supported our ideas with an additional study at the Institute for Technical Logistics. Our results regarding to this study are shared as suggestions within the paper .

## 3.1    Examining the replicability of LOCO dataset

The LOCO dataset has been the first open source dataset, which is prepared to be used in the intralogistics field. While this dataset was shared with the public in the GitHub repository, a paper was also published (Mayershofer et al., 2020). In this paper, it is possible to find information about the LOCO dataset. Mayershofer et al. shared so much information about the data collection process of the LOCO dataset and the evaluation results by using LOCO and Faster R-CNN algorithms were shared clearly. Especially, the information shared in the preparation of the dataset draws attention to the points to be considered for those who want to prepare such a dataset in the future.

There are remarkable points about video and photo capturing, which leads data acquisition process to high level. Firstly, images are captured from five different warehouses in real life scenarios while people and other vehicles are moving around. When images are examined in detail, it is realized that this brings high amount of diversity on the dataset. It was also important to visit several warehouses, as the color and model differences of the objects used in each warehouse increasing the variety in the dataset for real life scenarios.

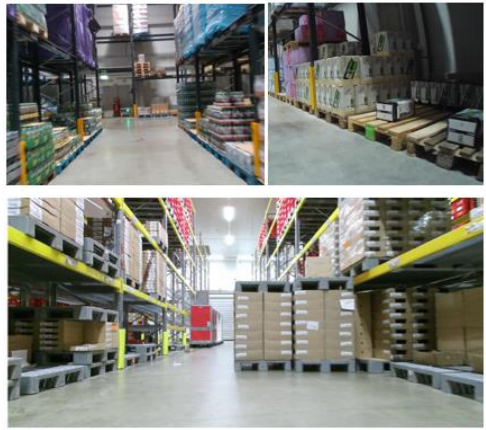

Figure 3: Blue, gray and different shades of brown colored pallets from different angles

In addition, the different size in the warehouses affect the way pallets are stored, and it also enables the collection of object images from a longer distance in large warehouses. Furthermore, the images are mostly collected within different brightness levels. Sometimes sun from the window is shining to the lens of the camera and sometimes images are captured in shady corners of the warehouse. All these different conditions enable mobile robots to be trained for different cases which might help to perform better in unexpected circumstances.

Images are collected by using five different cameras which are the commonly used in robotic applications. Each of the camera has high resolution values which differs between 1920 x 1080 and 1280 x 800 in pixel.

Table 2: Cameras in Use - Images were captured using different cameras

| Camera | Data | Resolution in pixel |
|--------|------|---------------------|
| MS Kinect v2 | RGBD | 1920 x 1080 |
| Intel Realsense D435 | RGBD | 1920 x 1080 |
| SJCAM SJ-400 | RGB | 1920 x 1080 |
| MS LifeCam HD-3000 | RGB | 1280 x 800 |
| Logitech C310 | RGB | 1280 x 800 |

Table 3: Number of Images According to Width and Height in Pixel

| Number of Images | Width | Height |
|------------------|-------|--------|
| 560 | 1280 | 800 |
| 617 | 1280 | 960 |
| 1460 | 1280 | 720 |
| 1655 | 1920 | 1080 |
| 805 | 1920 | 480 |

On the other hand, images are collected in five different pixel values as shown in the following table. The object detection algorithm, which is trained with distinct size of images, causes image and video tests with diverse sizes to give better results. In addition, even if randomization from the hyperparameters is set to none during training, images with five varied sizes are sufficient for diversity. At the same time, however, it should be kept in mind that large image sizes will mostly cause insufficient CUDA memory problem in GPUs.

Another important sensitivity in the LOCO dataset is personal privacy. As mentioned before, capturing images while workers working in the field with forklifts and pallet trucks are important to create a more realistic dataset. However, it causes some personal privacy concerns in the dataset. Within the study, it must be elicited that there is no invasion of a worker's privacy in taking and publishing pictures. Because of this reason, Mayershofer et al. deployed a neural network for automated face recognition so that workers' faces can be detected and made then unrecognizable, even before images are saved in the hard drive. Unfortunately, no further technical information was shared about how this process can be automatized: which neural network is used and how images are saved automatically in hardware. That is why, this section will not be giving so many technical insights to those who want to remove the human face from the dataset in the future.

Machine learning algorithms used in object detection applications have a very important role in obtaining high accuracy from the dataset. To evaluate LOCO dataset, Mayershofer et al. used the Darknet (GitHub, 2022b) and Detectron (GitHub, 2022c) framework to train three different neural networks: YOLOv4, YOLOv4-tiny and Faster R-CNN. The average precision (AP) metric at an intersection over union (IoU) of 0.50 was chosen as key performance indicator (Bochovskiy et al., 2020). Their results are available on the following table.

Table 4: Evaluation Results From the LOCO Dataset Publications

| Model | YOLOv4-608 | YOLOv4-tiny | Faster R-CNN |
|---|---|---|---|
| Dataset | LOCO | LOCO | LOCO |
| mAP@50 | 41% | 22.1% | 20.2% |
| Small load carrier | 27.7% | 18.1% | 28.3% |
| Pallet | 65.0% | 36.2% | 19.8% |
| Stillage | 53.1% | 31.3% | 37.6% |
| Forklift | 31.3% | 11.6% | 2.9% |
| Pallet truck | 28.1% | 13.3% | 12.5% |

Within our studies we preferred to work only with YOLO algorithms as both YOLOv4 and YOLOv4-tiny is compiled in Darknet framework and these two algorithms are already enough to make decisions during our critical evaluation. We reached significantly better results in both YOLOv4 and YOLOv4-tiny trainings, after 35 different training sessions. Training durations are varied from 12 to 36 hours according to hyperparameters and GPUs assigned by Google Colab services.

Table 5: Evaluation Results of Our Study

| Model | YOLOv4 | YOLOv4-tiny |
|---|---|---|
| Dataset | LOCO | LOCO |
| mAP@50 | 64.54% | 49.98% |
| Small load carrier | 62.33% | 50.18% |
| Pallet | 68.39% | 54.38% |
| Stillage | 67.31% | 51.26% |
| Forklift | 60.76% | 48.39% |
| Pallet truck | 58.01% | 42.2% |

Since the original publication does not contain a detailed description of the hyperparameters, it only can be speculated about the different results in terms of accuracy of the trainings. A potential reason could be the different data splitting approaches.

Splitting of data into training and test datasets is an important aspect of a machine learning tasks. Having a correct amount of distribution among train and validation datasets will bring significantly better results. Data splitting for training dataset can be given 60 percent of all data, if the dataset is relatively small which has got up to 1,000 labels. At the same time, however, it is also suggested that if there is high number of labels about 1 million, the dataset can even be split with 99% for training and 1% for validation, since there are already 10.000 labels in validation set which is still more than enough to validate a model.

In LOCO dataset there are exactly 151,427 labels in 5,097 images collected from five different warehouses and separated to five distinct subsets from each warehouse. They divide the five subsets as follows: subsets two, three and four serve as training purposes, while subsets one and five are used in evaluation. This results in ratio of 3/5 training and 2/5 validation split. However, within our study, the distribution of dataset is changed about 15-20 percent to make the training stronger. We included subset five to the training dataset and left the biggest subset one alone for validation purposes. Subset one has about 1,140 images and the split among images for train dataset increased to about 78 percent. This can be one of the reasons why our validation results are performing better results than evaluation results shared in LOCO publications.

On the other hand, object detection applications are generally used in real time scenarios. It is known that when it is related to being able to recognize and detect the objects in real time videos over 30fps, one stage neural networks such as YOLO algorithm performs better than two stage neural networks like Faster R-CNN algorithm as shown in Figure 2. However, no experience is commented for these two different types of algorithms in performance level within the publications from LOCO dataset. On the other hand, in the LOCO dataset publications, it is said that a video is available online(https://github.com/tum-fml/loco) to see how the algorithm trained by LOCO dataset works in video. However, the related video cannot be found on the GitHub repository.

When all the conditions for the replicability of the LOCO dataset are examined, it can be said that a lot of information has been shared about the preparation of the dataset. Image collection techniques are clear and repeatable according to its publications. However, even though training of a dataset is always an important step in object detection applications, there is no detailed information shared how algorithms were tuned with various hyperparameters.

## 3.2    Strengths and weaknesses of the LOCO dataset

The availability of open datasets is crucial for the development and training of machine learning algorithms. Therefore, it is even more important to examine the capabilities of these datasets and identify potentials for improvement. This paper presents the first critical replication study with the LOCO dataset for object detection on logistics. It shows that the training based on LOCO can even achieve a higher accuracy than presented in the original publication, as also mentioned in the previous part. However, even with these significantly better results, there is still noticeable room for improvement of the LOCO dataset, while obvious strengths of dataset are substantial. That is why understanding of both strengths and weaknesses is important.

Obtaining object detection successfully is mostly dependent on the conditions of the environment and the object to be detected. The LOCO dataset gives most of the time impressive results in pallet, small load carrier and stillage detections, even though these objects are smaller than others and the way how they are stored in warehouses is so random and not well organized. There are some conditions, which make the object detection harder to be accomplished with a favorable result such as long distance or half appearance of the objects in the image frame. However, as it can be seen on the Figure 4, there are many pallets detected even from far away on the right hand side of the picture.

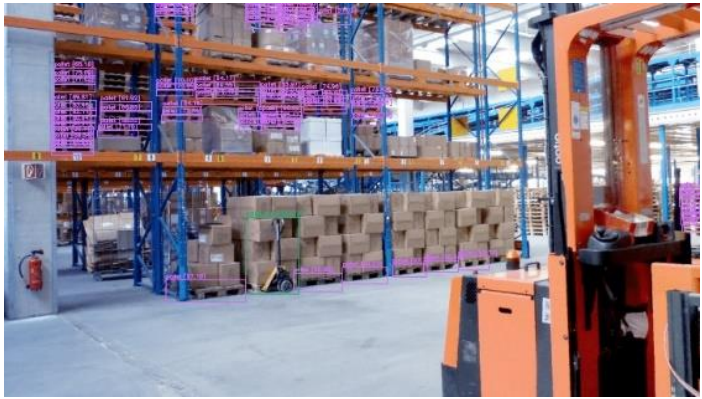

Figure 4: Detected pallet object from far away

In addition, stillage detection gives mostly positive impression even under difficult conditions. For example, two stillages draw attention in Figure 5. Even though both stillages have a half appearance on the image, they can be identified correctly by the algorithm.

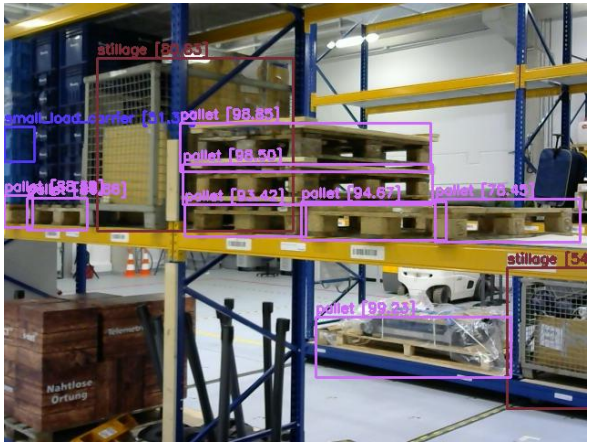

Figure 5: Efficient stillage detection

The reason why the LOCO dataset is very efficient in detecting these objects is revealed when the labels in the dataset are examined. Most of the LOCO dataset consists of pallets with approximately 120,445 labels. Then, it is followed by small load carrier with approximately 22,151 labels and stillages with 5,047 annotations. The fact that the storage work is handled with these three objects, which are the most common in warehouses, makes the high number of labels acceptable. This has resulted in the LOCO dataset giving mostly very strong results for these three types of object and specifically absolute good detection results for pallets.

Table 6: Number of Annotations per LOCO Classes

| Category | Number of Annotations |
|---|---|
| Pallet | 120,445 |
| Small load carrier | 22,151 |
| Stillage | 5,047 |
| Pallet truck | 2,827 |
| Forklift | 598 |

In addition, the number of forklift and pallet truck annotations in LOCO dataset is lower, when we saw the label amount for these classes at the beginning of the study. Later in the test phases of our study, it was noticed that the concern was justified, and some noticeable mistakes needed to be shared. In Figure 6, there is a forklift which is not so far away causing a confusion on the algorithm by recognizing both as forklift and pallet truck object at the same time. This situation clarifies how serious the current error among these LOCO classes is. Moreover, having such a mistake caused us to spend more time on forklifts and pallet trucks later in the study and the possible reasons are investigated with more images in the following phases.

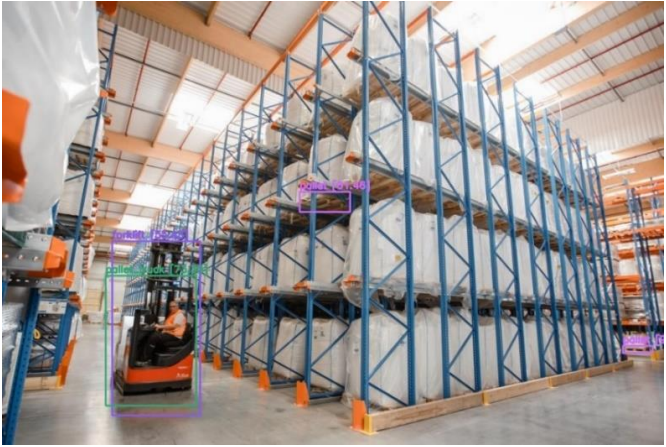

Figure 6: Confusion in forklift detection

The following group of pictures are from a warehouse which only includes all the LOCO objects to be detected. The image in Figure 7 is taken in relatively close distance to the objects than the image in Figure 8. In Figure 7 most of pallets, small load carriers and also stillages are detected correctly. However, even though pallet trucks and forklifts are in the same distance and angle, while pallet truck is detected with about 73 percent and this can be accepted as decent performance, the forklift is not detected with our algorithm. On the other hand, Figure 7 is only couple of meters more far away than Figure 8. There are question marks raised again for the white forklift, since it is detected as pallet truck which is a confusion of the algorithm.

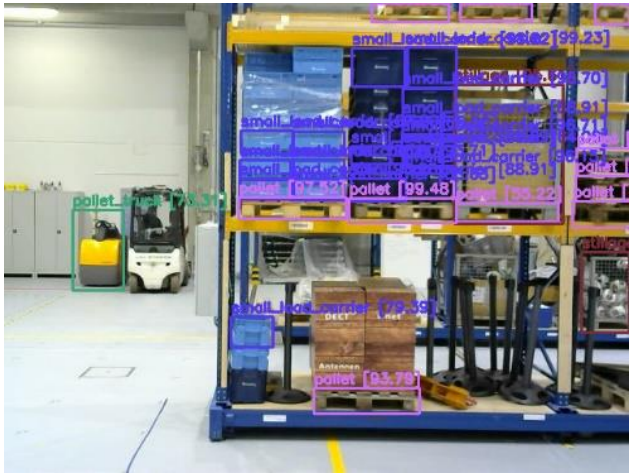

Figure 7: LOCO objects in detection test - close

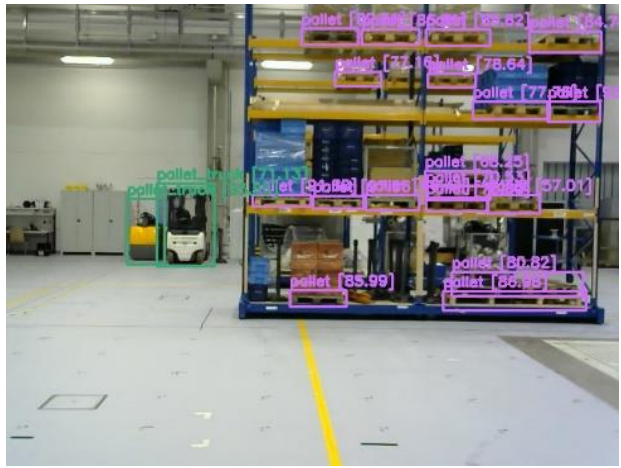

Figure 8: LOCO objects in detection test - away

Another factor that makes it difficult to detect forklifts is that various forklift types with very different functionalities have been developed and used in warehouses in the last decades. The total number of forklifts in the LOCO dataset is around 598 which is already a limited number for forklift annotations in the LOCO dataset compared to other classes. As it is explained previously, there are already some obvious problems while detecting traditional forklifts. On the other hand, the LOCO dataset should be considered one more time for innovative forklift vehicles developed with today's technology for extensive needs in warehouses. For instance, the following images show the test results from warehouses with relatively innovative forklifts. One of the forklifts is not detected, while others are recognized as pallet truck by the YOLO algorithm.

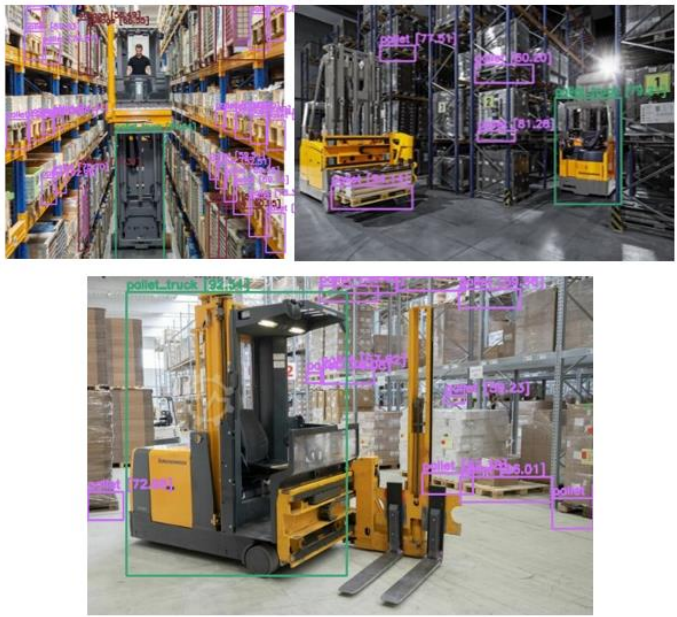

Figure 9: Innovative forklifts detected as pallet trucks

Moreover, while working with the LOCO dataset, it was noticed that annotated pallet trucks differ significantly from each other. Some pallet trucks are electric, some are non-electric and sometimes even fully manual. In addition, in some models, the operator can get on the pallet truck and move inside the warehouse with the vehicle, as they are drivable. However, the fact that all of them are defined as a single class and the number of annotations is obviously not very sufficient for such a diverse class. This causes the model not to learn properly during training. Therefore, the high confusion in the algorithm is noticeable during the test phase.

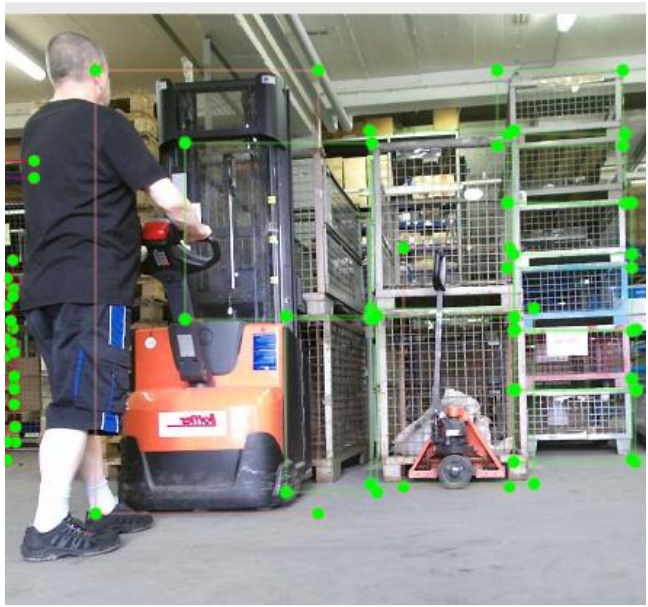

Figure 10: Electric powered and manual pallet trucks having different appearances

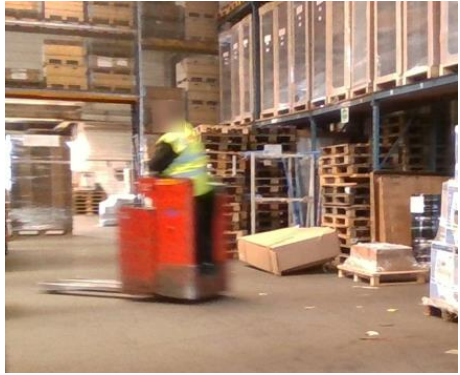

Figure 11: Operators can get onto some of pallet trucks

To be able to make the pallet truck class more precise, it could be suggested to separate it into three different classes and each of them should have a higher number of annotations within their classes. The suggested names for each of these pallet truck types could be called as: electric pallet truck, manual pallet truck, and drivable pallet truck.

### 3.2.1   Additional Study with Forklift and Pallet Trucks

Due to the unsatisfactory results of the LOCO dataset in forklift and pallet truck detection, we decided to create our own dataset with pallet trucks and forklifts. The goal was to experience by ourselves how to create a valuable dataset for object detection. The dataset is created and used as completely different study than the LOCO dataset. These two datasets are not combined anytime during our studies.

Data collection was carried out by using forklifts and electronic pallet trucks in the laboratory facilities of the Institute for Technical Logistics at the Hamburg University of Technology. A total number of 621 different images were collected from these two different vehicles. While collecting the images, operators and pallets were used as figurants to increase the suitability of dataset for the real life scenarios. The images of the vehicles were taken from different angles, heights and distances, thus increasing the diversity of the dataset. Some photos included in the dataset can be seen from the following Figure 12 and Figure 13.

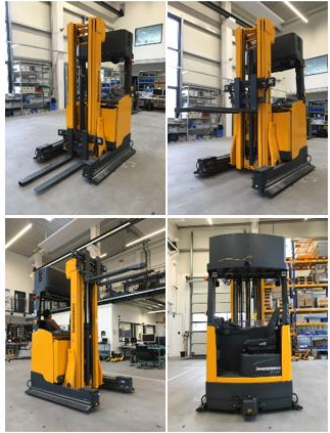

Figure 12: Image samples from a single forklift

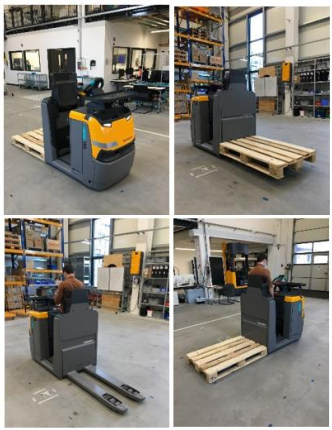

Figure 13: Image samples from a single pallet truck

At the training phase of our additional study, the YOLOv4 algorithm was trained with this dataset, which only includes forklift and pallet truck images. A high number of tests wase carried out to evaluate the algorithm.

This dataset performs well for various types of forklifts, see the following Figure 13. These forklifts could not be detected by the algorithm trained with the LOCO dataset. However, after training YOLOv4 algorithm with our own images, the algorithm was able to detect these forklifts without problem. We made a small examination at the end of this part for the reasons why our dataset gives very good results in such forklift types, while the algorithm with LOCO dataset fails most of the time.

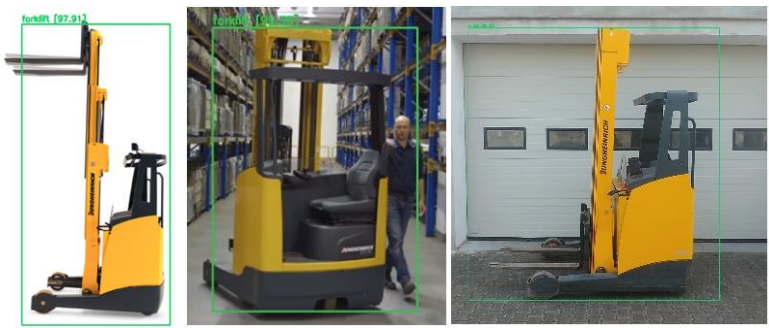

Figure 14: Forklift detection results with good impression

On the other hand, our dataset does not perform well under various conditions. Especially when there are multiple forklifts in the image and if some of the forklifts are far away, the detection results are so unsatisfactory. Some unacceptable errors were received during the test phase as it can be seen on Figure 14. The detections from algorithm trained by LOCO dataset were still not so good but still better than the test results from our additional study.

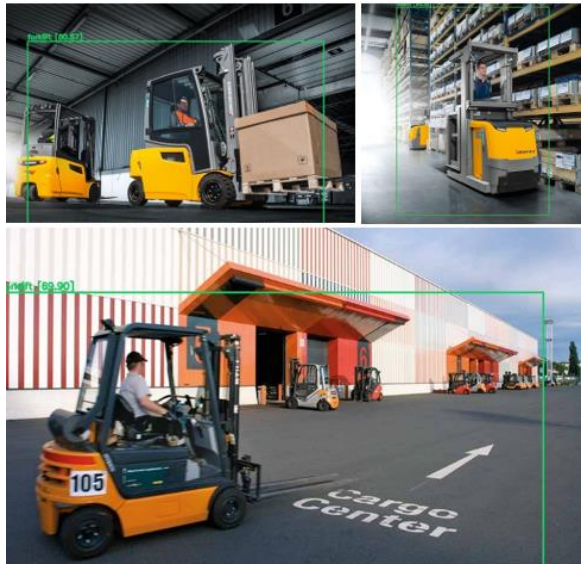

Figure 15: Significant mistakes in forklift detection

To be able to give some more general insights to researcher's studies in data collection, we would like to explain clearly what was missing in our dataset from our additional study so that these suggestions can help to make their data much accurate. There are several points to highlight as general suggestions for data preparation:

• An artificial area was created in the middle of the lab to collect images of vehicles. Moreover, the collected images would have been performing better, if they had been collected from regular working atmosphere in real scenarios.

• Our images are only focused in one forklift in each image. That is why, after the training the algorithm with our dataset, the algorithm was not aware of being able to detect multiple objects in the image.

• Forklift and pallet trucks are completely visible in our images. That is why, the algorithm does not perform well, when there is obstacle in front of the object.

- Forklifts and pallet trucks are mostly in the same distance range between two to four meters. That makes it harder to detect forklifts, which are more far away on the image.

- While creating our dataset, only one type of forklift and pallet trucks is used. When the algorithm is tested with different types of forklifts and pallet trucks, it will not give good results.

- Both of our forklifts and pallet trucks have yellow colour. That decreases the efficiency of the detection, even we test the algorithm with the same shape of vehicle but different colour of forklift.

This additional study let us to understand better about how to prepare a dataset for object detection applications. During data preparation, it is important to pay attention some main points which affect the diversity of the dataset respect to objects' variety with colour and types, environment conditions and object positions according to distance and angle between camera and object.

To sum up, after this part of our study, it can be interpreted better the strength and weaknesses of the LOCO dataset. The LOCO dataset gives outstanding detection performance for detecting pallets, small load carriers and stillages compared to detection of pallet trucks and forklifts. Our first suggestion is that LOCO dataset would give much better results for pallet trucks, if this class is extended as three different classes as explained before. Secondly, more forklift and pallet truck annotations are required than currently existing number of annotations in the LOCO dataset.

## 3.3    Data augmentation of the LOCO dataset

Data augmentation is a useful technique to enhance the performance of algorithms by increasing the amount of data with artificially modified copies of already existing images. Data augmentation is the preferred method in computer vision applications. In our study, we created copies of images artificially and included them into the LOCO dataset so that we could increase the size of overall dataset. There are many libraries and methods that can be used to create artificial copies for data augmentation. The imgaug library (imgaug — imgaug 0.4.0 documentation, 2022) provides so much flexibility for machine learning applications. It gives wide range of augmentation techniques to execute them in CPU cores. Moreover, there are also many methods included in imgaug library such as rotating, scaling, translating, flipping, adding blurry effect, changing brightness etc. In our augmentation study, we focused for rotation, scaling and translating the images in both x and z axis. Our augmentation study includes 3 different augmentation methods respectively: rotation, scaling and translating of each image exists in LOCO dataset. First, we randomly rotated each image between -10 and 10 degrees. Secondly, we decreased the size of those images between 80% to 90% of its original size. Lastly, we translated augmented frame randomly in both x and y directions about 5% of its augmented frame size. There are relative examples in Figure 15.

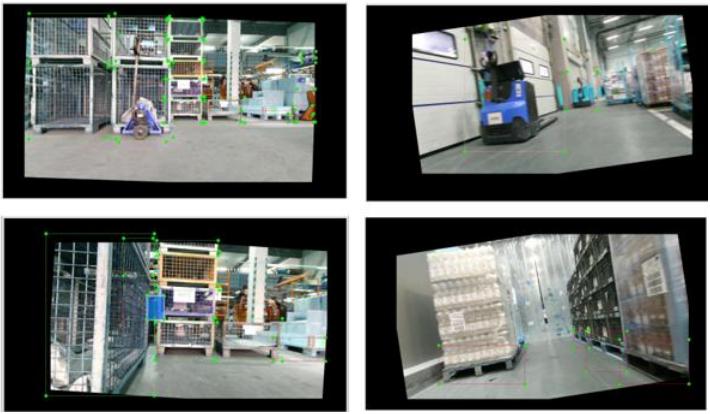

Figure 16: Augmented image samples

Since data augmentation is one of the basic methods to increase the accuracy in computer vision applications, we wanted to experience the effect of our own augmentation data which is later combined with LOCO dataset and trained both together in YOLOv4. Our result is only one percent better mAP compared to the evaluation results with the LOCO dataset. mAP@50 64.79 percent results for the validation results are the best results we have, as it can be visible in the Table 7.

To summarize, data augmentation by using rotation, scaling and translation improved our results a little bit less than we had expected. However, it is still in good order to experience improvements in the evaluation results. Since imgaug library offers many other augmentation use cases, in the future, some other variations of augmentations such as flipping, blurring, contrasting etc. can be examined with different combinations to improve results in the LOCO dataset.

Table 7: Evaluation Results of Our Study After Augmented Data Included

| Model | YOLOv4 |
| --- | --- |
| Dataset | LOCO & Augmented Data |
| Map@50 | 64.79% |
| Small load carrier | 62.21% |
| Pallet | 66.91% |
| Stillage | 71.03% |
| Forklift | 58.29% |
| Pallet truck | 59.24% |

# 4    Conclusions and future work

We presented our study about evaluating the LOCO dataset, which includes 5,097 annotated images with 151,427 labels from five different classes of objects: pallet, small load carrier, stillage, forklift and pallet truck. We concluded our research objective within three main results. Firstly, we are determined that the LOCO dataset is possible to replicate the findings according to its publications. There are significant points pointed out clearly about image collection techniques in object detection processes. However, there are several information missing about how the dataset was trained and with which hyperparameters three different object detection algorithms are tuned. Secondly, we examined the strengths and weaknesses of the LOCO dataset. We realized that YOLOv4 algorithm after training by the LOCO dataset gives impressive results, while detecting pallet, small load carrier and stillage. On the other hand, in the further stage of our study, distinctive errors are noticed about pallet truck and forklift detections. The algorithm was most of the time not detecting forklifts and pallet trucks on the test images or sometimes algorithm was confusing pallet truck and forklift object classes with each other. There are two root causes recognizable for poor detection in these classes. First of them is the limited annotation numbers for both forklifts and pallet trucks. Second of them is the superficial classification for pallet trucks. We suggested within our study that pallet trucks could be classified in three different classes because of its physical appearance and functionalities. Thirdly, as last research question, we investigated the data augmentation effect for the LOCO dataset. We augmented all the images in the LOCO dataset by using rotation, scaling and translation functions in imgaug library. After combining augmented images to the LOCO dataset, our performance indicator, mAP@50, is increased about one percent in evaluations.

After our critical evaluations, we summarize of three main suggestions to be able to reach state of art object detection in intralogistics by using the LOCO dataset. Firstly, the LOCO dataset must include a greater number of annotations for forklifts and pallet trucks to increase its detection capability. Secondly, while increasing the number of annotations, the dataset should be fed by up to date innovative variations of these vehicles according to what is used in intralogistics sector currently. Lastly, there should be an expansion while classifying the pallet truck objects. Pallet truck classes can be separated as three groups and these classes can be called as: electric pallet truck, manual pallet truck, and drivable pallet truck.

Within our study, we examined the LOCO dataset and its publications according to object detection requirements in intralogistics. After finding answers to our research questions, we briefly explained what can be done to improve the LOCO dataset in the future along with our recommendations. In addition, we hope that the findings we have explained will help the future studies in the field of computer vision while preparing the dataset. We wanted to contribute to the work for creating a more advanced and reliable dataset in the field of intralogistics within this manner.